\pdfoutput=1
\documentclass[lettersize,journal]{IEEEtran}
\usepackage{booktabs}       
\usepackage{amsmath,amsfonts}
\usepackage{algorithmic}
\usepackage{algorithm}
\usepackage{array}
\usepackage[caption=false,font=normalsize,labelfont=sf,textfont=sf]{subfig}
\usepackage{textcomp}
\usepackage{stfloats}
\usepackage{url}
\usepackage{verbatim}
\usepackage{graphicx}
\usepackage{cite}
\usepackage{bm}
\usepackage{multirow}
\begin{document}

\title{Temporal-adaptive Weight Quantization for Spiking Neural Networks}

\author{Han Zhang, Qingyan Meng, Jiaqi Wang, Baiyu Chen, Zhengyu Ma, Xiaopeng Fan~\IEEEmembership{Senior Member,~IEEE}
\thanks{This work was supported by National Science and Technology Innovation 2030 Major Project (No. 2025ZD0215501). This work was also supported in part by the National Natural Science Foundation of China (NSFC) under grant U22B2035. (Corresponding authors: Zhengyu Ma, Xiaopeng Fan.)}
\thanks{Han Zhang and Xiaopeng Fan are with the Faculty of Computing, Harbin Institute of Technology, Harbin 150001, China, and with Pengcheng Laboratory, Nanshan, Shenzhen 518000, China. Xiaopeng Fan is also with Suzhou Research Institute, Harbin Institute of Technology, Suzhou 215104, China (e-mail: 23b303002@stu.hit.edu.cn, fxp@hit.edu.cn).}
\thanks{Jiaqi Wang is with the Institute of Computing and Intelligence (ICI), Harbin Institute of Technology, Shenzhen 518055, China, and with Pengcheng Laboratory, Nanshan, Shenzhen 518000, China.}
\thanks{Baiyu Chen is with Institute of Automation, Chinese Academy of Sciences, Beijing 100190, China, and with Pengcheng Laboratory, Nanshan, Shenzhen 518000, China.}
\thanks{Qingyan Meng and Zhengyu Ma are with Pengcheng Laboratory, Nanshan, Shenzhen 518000, China (e-mail: mengqy@pcl.ac.cn, mazhy@pcl.ac.cn).}
\thanks{The code is available at https://github.com/ZhangHanN1/TaWQ}
}

\markboth{Journal of \LaTeX\ Class Files,~Vol.~xx, No.~x, October~2025}%
{Zhang \MakeLowercase{\textit{et al.}}: Temporal-adaptive Weight Quantization for Spiking Neural Networks}

\IEEEpubid{0000--0000/00\$00.00~\copyright~2021 IEEE}

\maketitle

\begin{abstract}
Weight quantization in spiking neural networks (SNNs) could further reduce energy consumption. However, quantizing weights without sacrificing accuracy remains challenging. In this study, inspired by astrocyte-mediated synaptic modulation in the biological nervous systems, we propose Temporal-adaptive Weight Quantization (TaWQ), which incorporates weight quantization with temporal dynamics to adaptively allocate ultra-low-bit weights along the temporal dimension. Extensive experiments on static (e.g., ImageNet) and neuromorphic (e.g., CIFAR10-DVS) datasets demonstrate that our TaWQ maintains high energy efficiency (4.12M, 0.63mJ) while incurring a negligible quantization loss of only 0.22\% on ImageNet.
\end{abstract}


\section{Introduction}
Inspired by biological nervous systems, spiking neural networks (SNNs) are regarded as the third-generation neural networks \cite{maass1997networks}. Their event-driven nature, which transmits information through binary spikes \cite{roy2019towards, scalingSDT}, enables accumulation (AC) operations to substitute multiply-accumulate (MAC) in the neural network. This paradigm shift in computation results in low energy consumption, making SNNs conducive to being deployed on resource-constrained devices.

Although SNNs already improve energy efficiency relative to artificial neural networks, they still fall short of the remarkable efficiency achieved by biological nervous systems. For instance, the human brain sustains about $8.6\times 10^6$ neurons and over $1\times 10^{14}$ synapses while operating at approximately 20 watts \cite{humanbrain20w, attentionsnn, fastsnn}. In contrast, the energy efficiency of SNNs deployed on neuromorphic chips remains significantly inferior to this biological benchmark \cite{truenorth, tianjic, loihi2}. In the nervous system, a presynaptic spike triggers the release of discrete, neurotransmitter-filled vesicles into the synaptic cleft. Since each vesicle carries a fixed, quantal amount of neurotransmitter, the synaptic strength depends on the discrete number of vesicles released\cite{EDWARDS2007835, Distincttransmission, Alteredpropertiesof}. This intrinsic transmitter quantization provides a biological analog for weight quantization strategies in SNNs: substituting full-precision weights with ultra-low-bit ones that count discrete quanta \cite{q-snns}, yielding exponential energy savings.

Quantization of weights has been generally exploited in artificial neural networks (ANNs), where full-precision weights are quantized to no more than 8-bit, and in some cases, as low as 1-bit \cite{xnor-net-eccv2016, bitnet}. In contrast, weight quantization of SNNs is only beginning to emerge. There are only a few related studies demonstrating that coupling low-bit weights with event-driven spikes can drastically cut energy consumption: the QSD-Transformer \cite{quantizedspikedriventransformer} quantizes weights to 4-bit precision, while Q-SNNs \cite{q-snns}, BESTformer \cite{binaryeventdrivenspikingtransformer}, and AGMM \cite{AGMM} employ 1-bit quantization. Nevertheless, weight quantization in SNNs without sacrificing accuracy remains a significant challenge, underscoring the need for biologically grounded strategies that narrow the efficiency gap with the nervous systems.

\begin{figure*}[!t]
    \centering
    \includegraphics[width=.8\textwidth]{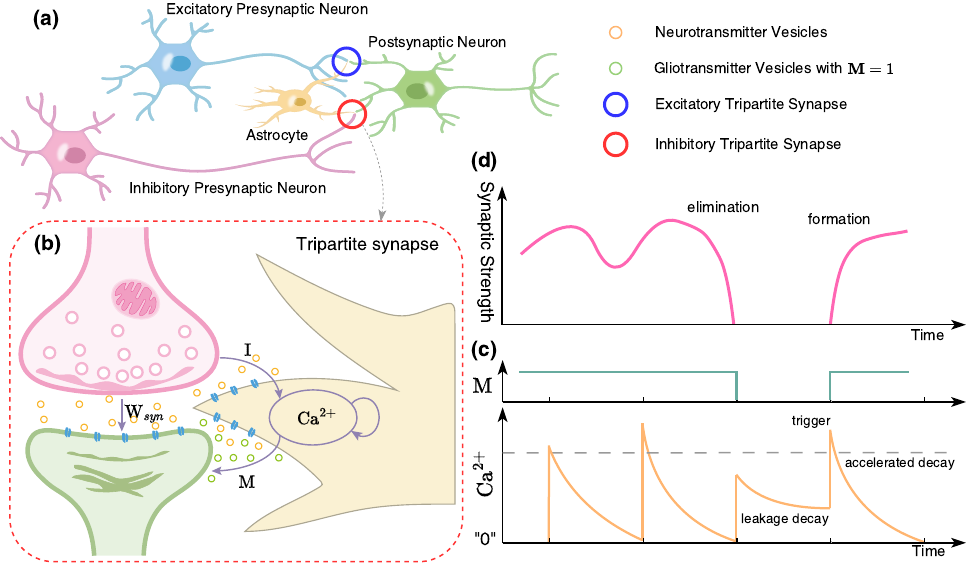}
    \vspace{-0.35cm}
    \caption{Schematic illustration of tripartite synapses. (a) An excitatory or inhibitory presynaptic neuron and a postsynaptic neuron, together with an astrocyte, form a tripartite synapse. (b) Schematic of the tripartite synapse structure, $\mathbf{W}_{syn}$ is the synaptic strength, $\mathbf{I}$ represents the stimulus received by astrocytes, $\mathbf{Ca}^{2+}$ denotes the calcium concentration, $\mathbf{M}$ is a symbol (not spike) designating whether astrocytes modulate synapses. (c) Calcium dynamics curve, triggering the $\mathbf{M}$ upon exceeding the threshold. (d) The synaptic strength varies over time under the modulation of astrocytes.}
    \label{fig: astrocyte_abcd}
    \vspace{-0.45cm}
\end{figure*}

Astrocytes are widely distributed in the biological nervous system, forming tripartite synapses with excitatory or inhibitory presynaptic and postsynaptic neurons \cite{tripartitesynapse}, and exhibit the ability to modulate synaptic strength across time \cite{rolesofreactiveastrocytes, astrocyticinhibition}, eliminate synapses \cite{astrocytesphagocytose, functionastrocytes}, and facilitate synapse formation \cite{astrocyteglypicans46}, which critically depend on intracellular calcium concentration oscillations.
As illustrated in Fig. \ref{fig: astrocyte_abcd}(d), we abstract astrocyte-mediated synaptic modulation into three principles: i) Under the modulation of astrocytes, synaptic strengths vary across time with invariant signs; ii) Astrocytes can eliminate some synapses, that is, convert excitatory or inhibitory synapses to asynaptic state; iii) Astrocytes can secrete thrombospondin and Hevin to facilitate the synapse formation, that is, convert the asynaptic state to excitatory or inhibitory synapses. The asynaptic state occurs either after the elimination of synapses or prior to synapse formation.

\IEEEpubidadjcol

Inspired by the tripartite synapse, we propose a novel quantization method termed Temporal-adaptive Weight Quantization (TaWQ), which integrates mechanisms with temporal dynamics into the quantization process, enabling weights to adopt distinct values across timesteps, so as to mitigate the challenge of performance degradation in weight quantization. This emulates the role of astrocytes in the modulation of synaptic strength by following the three abstract principles. This method constrains synaptic connectivity to three discrete states: excitatory synapses, asynaptic state, and inhibitory synapses, thereby converting full-precision floating-point weights into 1.58-bit ternary values $\{+1,0,-1\}$. 

Our main contributions are as follows. (1) First, we focus on the intracellular calcium concentration oscillations within astrocytes, extracting and modeling key characteristics of calcium dynamics, as astrocytic function is critically dependent on fluctuations in calcium concentration. (2) Based on the calcium dynamics model, we develop the Temporal-adaptive Weight Quantization (TaWQ) method, which integrates temporal dynamics into the weight quantization. Without additional trainable parameters, TaWQ quantizes full-precision floating-point weights into time-varying 1.58-bit ternary values $\{+1,0,-1\}$, corresponding to three synaptic states: excitatory, asynaptic, and inhibitory. (3) Extensive experiments on both static and neuromorphic datasets demonstrate that our TaWQ narrows the accuracy gap between full-precision and ultra-low-bit SNNs while enhancing their intrinsic energy advantages.

\section{Related Works}

\textbf{Ultra-Low-Bit Weight Quantization in Artificial Neural Networks.} 
XNOR-Net \cite{xnor-net-eccv2016} is a classical ultra-low-bit quantization method that compresses full-precision weights and activations into $\{+1, -1\}$. The dot product between binary vectors is implemented via XNOR-bitcounting operations, while the introduction of scaling factors mitigates accuracy degradation caused by quantization. DoReFa-Net \cite{dorefa-net} pioneers the comprehensive ultra-low-bit quantization of weights (1-bit), activations (2-bit), and gradients (6-bit), significantly reducing overhead and enhancing training efficiency. Bi-Real Net \cite{birealnet} enhances the network's representational capacity by introducing an optimized residual structure and a refined parameter updating algorithm. ReActNet \cite{reactnet} proposes generalized activation functions and a distributional loss to narrow the performance gap to full-precision baselines. Although these methods achieve remarkable hardware efficiency, they incur a noticeable performance degradation. BitNet \cite{bitnet} and its ternary extension BitNet-1.58 \cite{bitnet-1.58} demonstrate that ultra-low-bit weight quantization can preserve high performance while yielding substantial energy savings. However, their activations remain at 8-bit, leaving considerable potential for further efficiency refinement when both weights and activations are pushed into the binary-ternary regime.

\textbf{Ultra-Low-Bit Weight Quantization in Spiking Neural Networks.} 
Event-driven SNNs naturally achieve low-power inference since information is transmitted only through binary spikes, utilizing sparse and additive synaptic operations instead of multiply-accumulate operations. Weight quantization could further amplify these efficiency advantages. QP-SNNs \cite{qp-snns} employs a weight rescaling strategy that efficiently utilizes bit-width to enhance uniform quantization methods, thereby strengthening the model's representational capability, and the accuracy on ImageNet is only 61.36\% with 8-bit weights. QSD-Transformer \cite{quantizedspikedriventransformer} employs an information-enhanced LIF neuron and fine-grained distillation to mitigate performance degradation. The accuracy on ImageNet is 80.3\%. However, 4-bit quantization retains potential for further bit-width reduction, as some works have implemented 1-bit weight quantization. The Q-SNNs \cite{q-snns} binarize synaptic weights into $\{+1, -1\}$, and introduce a loss function to constrain neuronal firing rates approaching 0.5, and further reduce memory footprint by implementing 8-bit quantization on membrane potentials, but it exhibits noticeable performance degradation of 0.82\% and 0.75\% on the CIFAR-10 and CIFAR-100 datasets, respectively. AGMM \cite{AGMM} proposes an adaptive gradient modulation to ensure the performance after quantization, with 64.67\% accuracy on ImageNet. In addition to the network weights, BESTformer \cite{binaryeventdrivenspikingtransformer} further quantizes the self-attention map to 1-bit, with the resultant performance degradation being compensated by the proposed Coupled Information Enhancement strategy, the accuracy is 63.46\% on ImageNet. While existing methods attain efficient computation and model compression via integrating SNNs with ultra-low-bit quantization, they still require further optimization to narrow the performance gap with full-precision networks.

\section{Method} \label{Method}

\begin{figure*}[!t]
    \centering
    \includegraphics[width=.925\textwidth]{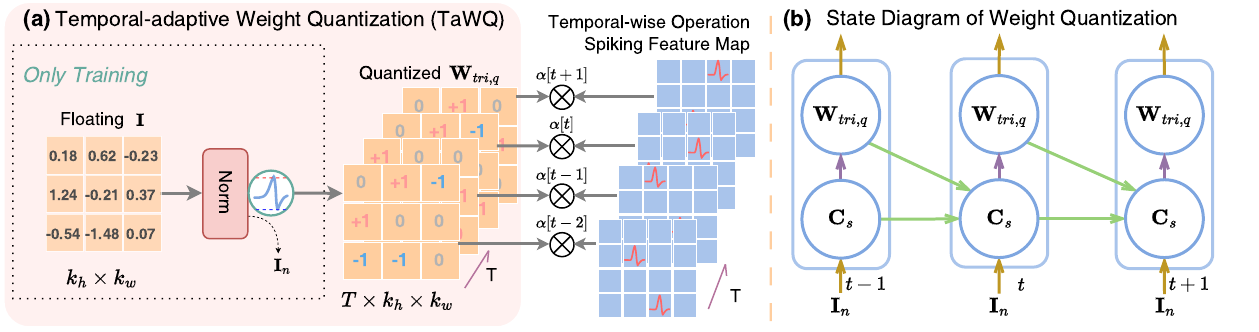}
    \vspace{-0.3cm}
    \caption{Schematic illustration of TaWQ. (a) Weights are quantized into time-varying 1.58-bit values \{+1, 0, -1\}, followed by temporal-wise operation. (b) The state diagram in the weight quantization process, $\mathbf{I}_{n}$, $\mathbf{C}_s$, and $\mathbf{W}_{tri,q}$ are the normalized stimulus, intermediate variable, and quantized weight, respectively.}
    \label{fig: tawq_ab}
    \vspace{-0.4cm}
\end{figure*}

\subsection{Calcium Dynamics in Astrocytes}\label{subsec:AstrocytesCalcium}
Synaptic states in biological nervous systems persistently evolve throughout development, and even in mature systems, they remain dynamic to enable complex functionalities. This phenomenon underscores the necessity of integrating SNNs with temporally evolving synaptic states, such integration significantly enhances the representational capacity of SNNs, thus mirroring the biological characteristic wherein synaptic dynamics facilitate sophisticated neural functions. Astrocytes are widely distributed throughout biological nervous systems and actively participate in the modulation of synapses. The astrocytic modulation of synapses represents a critical mechanism for enabling changes in synaptic states. As illustrated in Fig. \ref{fig: astrocyte_abcd}(b), this modulation process relies on intracellular calcium concentration oscillations within astrocytes. 

These calcium dynamics exhibit three key characteristics: 1) calcium concentration accumulates over time \cite{ca-accumulate}, 2) calcium concentration undergoes leakage decay \cite{ca-leak}, and 3) accelerated decay occurs upon exceeding a calcium concentration threshold \cite{ca-threshold}. These principles form the foundation of our motivation. We deliberately model astrocytic calcium dynamics by collapsing the temporal aspect, expressed mathematically as follows:
\begin{equation}\label{camodel}
\begin{cases}
\mathbf{Ca}^{2+}\left[t+1\right]=\kappa_1\cdot\mathbf{Ca}^{2+}[t](1-\mathbf{M}[t])+\kappa_2\cdot\lvert\mathbf{I}\rvert,\\
\mathbf{M}[t+1]=\mathcal{H}(\mathbf{Ca}^{2+}[t+1],C_{th}),\\
\mathbf{W}_{tri}[t+1]=\mathcal{F}(\mathbf{M}[t+1],\mathbf{W}_{syn}),
\end{cases}
\end{equation}
where $\kappa_1$ and $\kappa_2$ are scaling factors, $\mathbf{Ca}^{2+}[t]$ denotes the calcium concentration at timestep $t$, $\mathbf{M}[t]$ is a symbol designating whether astrocytes modulate synapses at timestep $t$, not a spike. $\mathbf{I}$ represents the stimulus received by astrocytes. A non-negative value for $\mathbf{Ca}^{2+}[t]$ is guaranteed by taking the absolute value of $\mathbf{I}$. $\mathcal{H}(\mathbf{Ca}^{2+},C_{th})$ is the Heaviside step function, which equals 1 if $\mathbf{Ca}^{2+} \geq C_{th}$, otherwise equals 0. And $\mathbf{W}_{syn}$ denotes the synaptic strength, $\mathbf{W}_{tri}$ represents the astrocyte-modulated synaptic strength, and 
$\mathcal{F}(\cdot)$ signifies the modulation function, as illustrated in Fig. \ref{fig: astrocyte_abcd}(a).

Eq. (\ref{camodel}) models key characteristics of calcium dynamics. The calcium concentration accumulates over time, with the scaling factor $\kappa_1$ controlling the rate of leakage decay, and when $\mathbf{M}=1$, it induces an accelerated decrease, as illustrated in Fig. \ref{fig: astrocyte_abcd}(c), in which the label "0" denotes the calcium concentration in the absence of stimulus. Although the calcium concentration is not absolutely constant without stimulus, we simplify the model by setting it to 0. This means that when the stimulus $\mathbf{I}=0$, the resulting  calcium concentration $\mathbf{Ca}^{2+}[t]=0$.

\subsection{Temporal-adaptive Weight Quantization}\label{subsec:Temporal-adaptiveWeightQuantization}
To narrow the performance gap induced by weight quantization, we draw inspiration from tripartite synapses and introduce dynamics into the quantization based on the calcium dynamics model described in Eq. (\ref{camodel}), proposing a novel SNNs-specific weight quantization method termed Temporal-adaptive Weight Quantization (TaWQ). The quantized weights exhibit temporal variability, adaptively switching between synaptic (excitatory or inhibitory) and asynaptic states. This behavior aligns with astrocyte-mediated synaptic modulation.

\textbf{1.58-bit TaWQ.}
We first normalize the stimulus $\mathbf{I}$ to ensure that they are subject to a distribution $\mathcal{N}(0,1)$, which follows \cite{weightnorm}, and it can be formulated as:
\begin{equation}
    \mathbf{I}_n=\frac{\mathbf{I}-\mu_{I}}{\sqrt[2]{\sigma_{I}^2+\epsilon}}.
\end{equation}

The stimulus $\mathbf{I}$ and the weight $\mathbf{W}_{syn}$ share the same shape, indicating that each synapse receives its respective stimulus input. $\mathbf{W}_{syn},\mathbf{I}\in \mathrm{R}^{C_o\times C_i\times k_h\times k_w}$ and $\mathbf{I}_n\in \mathrm{R}^{C_o\times C_i\times k_h\times k_w}$ denote the stimulus before and after normalization, where $C_o$ and $C_i$ represent the number of output and input channels in convolution layers (Conv), $k_h$ and $k_w$ denote the height and width of the convolution kernel, and $\mu_{I}$ and $\sigma_{I}$ denote the mean and standard deviation of $\mathbf{I}$, respectively. $\epsilon$ is an infinitesimal constant introduced to avoid division-by-zero scenarios. The stimulus generated by excitatory synapses should correspondingly induce excitation, and the same logic applies to inhibitory synapses. Therefore, the sign of $\mathbf{I}_n$ should be consistent with that of $\mathbf{W}_{syn}$, that is, $\mathrm{sign}(\mathbf{I}_n)=\mathrm{sign}(\mathbf{W}_{syn})$.

\begin{figure}[!t]
    \centering
    \vspace{-0.1cm}
    \includegraphics[width=.5\textwidth]{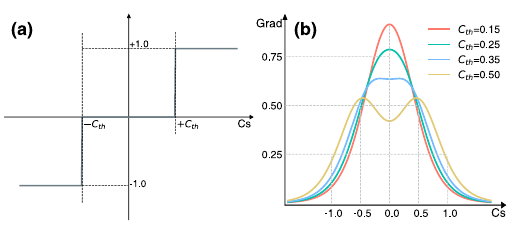}
    \vspace{-0.8cm}
    \caption{Curve of the quantization function and surrogate gradient, "Cs" on the horizontal axis is $\mathbf{C}_s$ in Eq. (\ref{singlebittawq}). (a) The quantization function converts floating-point values into 1.58-bit ternary values $\{+1, 0, -1\}$. 
    (b) Surrogate gradient under varying thresholds $C_{th}$.}
    \label{fig: TaWQ-step-SG-group}
    \vspace{-0.45cm}
\end{figure}

Subsequently, we define the $\mathcal{F}(\mathbf{M},\mathbf{W}_{syn})=\mathbf{M}\cdot\mathbf{W}_{syn}$. To focus solely on the presence or absence of synaptic connections rather than their specific strengths, the matrix $\mathbf{W}_{syn}$ is replaced with $\mathrm{sign}(\mathbf{W}_{syn})$, which effectively reduces the function to:
\begin{equation}
    \mathcal{F}(\mathbf{M},\mathbf{W}_{syn})=\mathbf{M}\cdot\mathrm{sign}(\mathbf{W}_{syn})=\mathbf{M}\cdot\mathrm{sign}(\mathbf{I}_{n}).
\end{equation}
Given the constraint $\kappa_1+\kappa_2=1$, if $\kappa_1=\lambda$, then $\kappa_2=1-\lambda$. The expression for qiantized $\mathbf{W}_{tri}$ can be rewritten as: 
\begin{equation}\label{camodel2}
\begin{cases}
\mathbf{Ca}^{2+}\left[t+1\right]=\lambda\cdot\mathbf{Ca}^{2+}[t](1-\mathbf{W}_{tri,q}[t])+(1-\lambda)\cdot\lvert\mathbf{I}_n\rvert,\\
\mathbf{W}_{tri,q}[t+1]=\mathcal{H}(\mathbf{Ca}^{2+}[t+1],C_{th})\cdot\mathrm{sign}(\mathbf{I}_{n}).
\end{cases}
\end{equation}
The $\mathcal{H}(\mathbf{Ca}^{2+},C_{th})\cdot\mathrm{sign}(\mathbf{I}_{n})$ is equivalent to the quantization function $\mathcal{S}(\mathbf{Ca}^{2+}\cdot\mathrm{sign}(\mathbf{I}_{n}),C_{th})$, which is defined as Eq. (\ref{firetawq}) and illustrated in Fig. \ref{fig: TaWQ-step-SG-group}(a):
\begin{equation}\label{firetawq}
    \mathcal{S}(\mathbf{Ca}^{2+}\cdot\mathrm{sign}(\mathbf{I}_{n}),C_{th})=\begin{cases}+1, &\mathbf{Ca}^{2+}\cdot\mathrm{sign}(\mathbf{I}_{n}) > +C_{th}, \\-1, &\mathbf{Ca}^{2+}\cdot\mathrm{sign}(\mathbf{I}_{n}) < -C_{th}, \\ 0,  &\text{else}.
\end{cases} 
\end{equation}

To distinguish the effects of excitatory/inhibitory stimulus on calcium concentration while simplifying the expression, we use intermediate variable $\mathbf{C}_s$ in the real domain to denote the $\mathbf{Ca}^{2+}\cdot\mathrm{sign}(\mathbf{I}_{n})$, thereby obtaining the expression for the 1.58-bit TaWQ as follows: 
\begin{equation}\label{singlebittawq}
\begin{cases}
\mathbf{C}_s\left[t+1\right]=\lambda\cdot\mathbf{C}_s[t](1-\lvert\mathbf{W}_{tri,q}[t]\rvert)+(1-\lambda)\cdot\mathbf{I}_{n},\\
\mathbf{W}_{tri,q}[t+1]=\mathcal{S}(\mathbf{C}_s[t+1],C_{th}),
\end{cases}
\end{equation}
The forward propagation state diagram of $\mathbf{C}_s$ is illustrated in Fig. \ref{fig: tawq_ab}(b). $\mathbf{W}_{tri,q}\in \mathrm{R}^{T\times C_o\times C_i\times k_h\times k_w}$ represents the quantized 1.58-bit weight, $T$ is timesteps, $\lambda=0.5$ serves as the scalar scaling coefficient, and $C_{th}$ indicates the quantization threshold. For Linear layers, both $k_h$ and $k_w$ are set to 1.

The TaWQ in SNNs quantizes weights $\mathbf{W}_{syn}$ into 1.58-bit time-varying $\mathbf{W}_{tri,q}$. Following the TaWQ, we subsequently introduce a temporal-wise scaling factor to compensate for performance loss caused by weight quantization. Let $\mathbf{X}_i\in\mathcal{R}^{T\times C_i\times H\times W}$ denote the input to the Quantized Conv Layer (Q-Conv), where $H$ and $W$ represent the height and width of $\mathbf{X}_i$, respectively. The quantized weights of Q-Conv are $\mathbf{W}_{tri,q}$. 
As illustrated in Fig. \ref{fig: tawq_ab}(a), the computation of the Q-Conv can be formulated as:
\begin{equation}\label{Q-Conv_single_step_equation}
    \mathbf{X}_o[t]=(\bm{\alpha}[t]\odot\mathbf{W}_{tri,q}[t])\otimes \mathbf{X}_i[t]=\bm{\alpha}[t]\odot(\mathbf{W}_{tri,q}[t]\otimes \mathbf{X}_i[t]).
\end{equation}
In the above expression, $\mathbf{X}_o\in\mathcal{R}^{T\times C_o\times H^{\prime}\times W^{\prime}}$ is output feature map of Q-Conv, $\otimes$ denotes matrix product, $\odot$ is an element-wise product, the parameter $\bm{\alpha}\in\mathcal{R}^{T\times C_o}$ is the temporal-wise scaling factor. After training, $\bm{\alpha}$ becomes fixed and can be folded into subsequent computational steps during inference. $\bm{\alpha}$ defined as:
\begin{equation}
    \bm{\alpha}[t,c]=\phi(\frac{1}{C_ik_hk_w}\sum\limits_{i=0}^{C_i-1}\sum\limits_{j=0}^{k_h-1}\sum\limits_{k=0}^{k_w-1}\lvert\mathbf{W}_{tri,q}[t,c,i,j,k]\rvert).
\end{equation}

The $\phi(\cdot)$ denotes the reciprocal function. For multiple timesteps, the computation of the Q-Conv is shown as follows:
\begin{equation}\label{Q-Conv:stack}
    \mathbf{X}_o=\mathrm{Stack}(\mathbf{X}_o[0],\mathbf{X}_o[1],\ldots,\mathbf{X}_o[T-1]),
\end{equation}

During the inference phase, the temporal-wise scaling factor $\bm{\alpha}$ of Q-Conv and the trained parameters of batch normalization (BN) can be jointly folded into the spiking neuron. If the LIF neurons are employed in SNNs, when the reset membrane potential is $0$, their charging process can be described as: 
\begin{equation}\label{folded_charging}
\begin{aligned}
    \mathbf{U}[t]=&(1-\frac{1}{\tau})\mathbf{U}[t-1]+\frac{1}{\tau}(\gamma\frac{\bm{\alpha}[t-1]\odot\mathbf{X}_q[t-1]-\mu}{\sqrt[2]{\sigma^2+\epsilon}}+\beta)\\
    =&(1-\frac{1}{\tau})\mathbf{U}[t-1]+\bm{\rho}[t-1]\odot\mathbf{X}_q[t-1]+\delta,
\end{aligned}
\end{equation}
\begin{equation}
    \bm{\rho}[t-1]=\frac{\gamma\bm{\alpha}[t-1]}{\tau\sqrt[2]{\sigma^2+\epsilon}},\quad \delta=\frac{1}{\tau}(\beta-\frac{\gamma\mu}{\sqrt[2]{\sigma^2+\epsilon}}).
\end{equation}
$\mathbf{U}[t]$ denotes the accumulated membrane potential, $\mathbf{X}_q[t]=\mathbf{W}_{tri,q}[t]\otimes \mathbf{X}_i[t]$, $\tau$ represents the membrane time constant, $\mu$ and $\sigma$ are the mean and standard deviation of $\bm{\alpha}\odot\mathbf{X}_q$, while $\gamma$ and $\beta$ correspond to the scaling and shift factors of BN.

\textbf{Backpropagation.}
Gradients of $\mathbf{I}$ are calculated as $\frac{\partial L}{\partial\mathbf{I}}=\frac{\partial L}{\partial\mathbf{I}_{n}}\frac{\partial \mathbf{I}_{n}}{\partial\mathbf{I}}$, and $\frac{\partial L}{\partial\mathbf{I}_{n}}$ is as follows:
\begin{equation}\label{LWn}
\begin{aligned}
\frac{\partial L}{\partial\mathbf{I}_{n}}=&\sum_{t=1}^T\frac{\partial L}{\partial\mathbf{W}_{tri,q}[t]}\frac{\partial\mathbf{W}_{tri,q}[t]}{\partial\mathbf{C}_s[t]}(\frac{\partial\mathbf{C}_s[t]}{\partial\mathbf{I}_{n}}\\
+&\sum_{j<t}\prod_{i=1}^{t-j}(\frac{\partial\mathbf{C}_s[t-i+1]}{\partial\mathbf{C}_s[t-i]}\\
+&\frac{\partial\mathbf{C}_s[t-i+1]}{\partial\mathbf{W}_{tri,q}[t-i]}\frac{\partial\mathbf{W}_{tri,q}[t-i]}{\partial\mathbf{C}_s[t-i]}\Bigg)\frac{\partial\mathbf{C}_s[j]}{\partial\mathbf{I}_{n}}\Bigg),
\end{aligned}
\end{equation}
where the term $\frac{\partial\mathbf{W}_{tri,q}[t]}{\partial\mathbf{C}_s[t]}$ denotes the derivative of $\mathcal{S}(\mathbf{C}_s,C_{th})$, which is non-differentiable and will be replaced by the surrogate gradient in Fig. \ref{fig: TaWQ-step-SG-group}(b), formulated as follows:
\begin{equation}
    \frac{\partial\mathbf{W}_{tri,q}}{\partial\mathbf{C}_s}=\frac{1}{2}(\theta^\prime(4(\mathbf{C}_s+C_{th}))+\theta^\prime(4(\mathbf{C}_s-C_{th}))).
\end{equation}
The value of $\frac{\partial\mathbf{W}_{tri,q}}{\partial\mathbf{C}_s}$ is composed of two surrogate gradient, where $\theta^\prime(4(\mathbf{C}_s-C_{th}))$ and $\theta^\prime(4(\mathbf{C}_s+C_{th}))$ are derivative of Sigmoid. The convergence analysis for Eq. (\ref{LWn}) is provided in the Supplementary Material.
The update rule for $\mathbf{I}$ is:
\begin{equation}\label{weightupdate}
    \mathbf{I}^+=\mathbf{I}^- - \eta\cdot\mathbf{G},
\end{equation}
where $\eta$ denotes the learning rate, and $\mathbf{G}$ represents a gradient-related term whose formulation varies depending on the optimizer employed.

TaWQ achieves dual optimization. On the one hand, it implements quantization by converting floating-point weights into 1.58-bit ternary values $\{+1, 0, -1\}$, corresponding to excitatory synapses, asynaptic state, and inhibitory synapses, respectively, further exploiting SNNs' inherent energy efficiency. On the other hand, we innovatively integrate temporal dynamics into the quantization, enabling time-varying synaptic strength in quantized SNNs, which significantly enriches the representational capacity of quantized SNNs.

\begin{table*}[t]
  \centering
  \caption{Results on the ImageNet dataset. "Power (mJ)" indicates energy consumption, and "Acc (\%)" is the top-1 accuracy. The Power '0.99(2.54\%)' denotes an energy consumption of 0.99mJ, which represents only 2.54\% of the full-precision counterpart. The accuracy of all comparison methods is sourced from their original research publications.}
    \begin{tabular}{p{9.22em}cccccc}
    \toprule
    \multicolumn{1}{l}{Method} & Architecture & Weight Bits  & TimeStep  & Size(M)  & Power(mJ) & Acc(\%) \\
    \midrule
    \multicolumn{1}{l}{XNOR-Net \cite{xnor-net-eccv2016}} & ResNet18 & 1   & 1     & -     & -     & 51.2 \\
    \multicolumn{1}{l}{Bi-Real Net \cite{birealnet}} & Bi-Real-18 & 1   & 1     & -     & -     & 56.4 \\
    \multicolumn{1}{l}{AdaBin \cite{adabin}} & ResNet18 & 1   & 1     & -     & -     & 66.4 \\
    \multicolumn{1}{l}{ReActNet \cite{reactnet}} & ReActNet-A & 1   & 1     & -     & -     & 69.4 \\
    \midrule
    \multicolumn{1}{l}{QP-SNN \cite{qp-snns}} & ResNet-18 & 8   & 4     & 13.28 & -     & 61.36 \\
    \multicolumn{1}{l}{BESTformer \cite{binaryeventdrivenspikingtransformer}} & BESTformer-8-512 & 1   & 4     & 5.57  & -     & 63.46 \\
    \multicolumn{1}{l}{AGMM \cite{AGMM}} & ResNet-18 & 1   & 4     & -     & -     & 64.67 \\
    \midrule
    \multirow{3}{*}{QSDTransformer \cite{quantizedspikedriventransformer}} & SDTransformer-v2-T & 4  & 4     & 1.8 & 2.5  & 77.5 \\
                                                  & SDTransformer-v2-M & 4  & 4     & 3.9 & 5.7  & 78.9 \\
                                                  & SDTransformer-v2-L & 4  & 4     & 6.8 & 8.7  & 80.3 \\
    \midrule
    \multicolumn{1}{l}{Spikformer \cite{zhou2023spikformer}} & Spikformer-8-384 & 32  & 4     & 16.81 & 7.73  & 70.24 \\
    \multicolumn{1}{l}{SDTransformer \cite{spikedriventransformer}} & SDTransformer-8-384 & 32  & 4     & 16.81 & 3.90   & 72.28 \\
    \midrule
    \multirow{3}{*}{Spikingformer-CML \cite{cml}} & Spikingformer-8-384 & 32  & 4     & 16.81 & 4.69  & 74.35 \\
                                                  & Spikingformer-8-512 & 32  & 4     & 29.68 & 7.46  & 76.54 \\
                                                  & Spikingformer-8-768 & 32  & 4     & 66.34 & 13.68  & 77.64 \\
    \multirow{3}{*}{\textbf{Spikingformer-TaWQ}} & Spikingformer-8-384 &1.58  & 4     & \textbf{1.25(7.44\%)} & \textbf{0.33(7.04\%)}  & \textbf{72.44(-1.91)} \\
                                                 & Spikingformer-8-512 & 1.58  & 4     & \textbf{2.03(6.84\%)} & \textbf{0.43(5.76\%)}  & \textbf{75.17(-1.37)} \\
                                                 & Spikingformer-8-768 & 1.58  & 4     & \textbf{4.12(6.21\%)} & \textbf{0.63(4.61\%)}  & \textbf{77.42(-0.22)} \\
    \midrule

    \multicolumn{1}{l}{QKFormer \cite{qkformer}} & QKFormer-10-768 & 32  & 4     & 64.96 & 38.91 & 84.22 \\
    \multicolumn{1}{l}{\textbf{QKFormer-TaWQ}} & QKFormer-10-768 & 1.58 & 4     & \textbf{4.05(6.23\%)}      & \textbf{0.99(2.54\%)}      & \textbf{82.94(-1.28)} \\
    
    \bottomrule
    \end{tabular}%
  \label{tab:imagenetresults}%
  \vspace{-0.35cm}
\end{table*}%

\section{Experiments} \label{Experiments}
We evaluate the proposed quantization method on static datasets (ImageNet \cite{imagenet}, CIFAR-10/100 \cite{cifar10and100}), neuromorphic datasets (CIFAR-10-DVS \cite{cifar10dvs}, DVS128-Gesture \cite{dvs128}), and the speech dataset (SHD \cite{SHD}, in Supplementary Material). And we provide the theoretical energy consumption calculation method for TaWQ in the Supplementary Material, along with the detailed process for estimating energy consumption that incorporates the overhead associated with reading/writing weights and feature maps on hardware. Detailed experimental setups are also provided in the Supplementary Material.

\subsection{Results on ImageNet Classification}

\textbf{Comparing with other SNNs.}
The results on ImageNet are summarized in Table \ref{tab:imagenetresults}. All models in the table employ 1-bit activations, except for QSD-Transformer. The Spikingformer-TaWQ models achieve compact model sizes of 1.25M, 2.03M, and 4.12M, while attaining top-1 accuracies of 72.44\%, 75.17\%, and 77.42\%, with corresponding energy consumptions of 0.33mJ, 0.43mJ, and 0.63mJ, respectively. The QKFormer-TaWQ achieves a model size of 4.05M, an accuracy of 82.94\%, and an energy consumption of only 0.99mJ.
\textbf{i) Full-precision SNNs.} 
The model size of Spikingformer-TaWQ is only 7.44\%, 6.84\%, and 6.21\% of its full-precision counterpart, Spikingformer-CML, while its energy consumption corresponds to merely 7.04\%, 5.76\%, and 4.61\% of the latter. Notably, the performance degradation is minor, with Spikingformer-TaWQ exhibiting an accuracy reduction of only 0.22\% in the "8-768" architecture. Furthermore, in the "8-384" architecture, Spikingformer-TaWQ delivers higher accuracy than both the full-precision Spikformer and SDTransformer. The model size of the QKFormer-TaWQ is only 6.23\% of its full-precision counterpart, QKFormer, with an energy consumption of merely 2.54\% and a minor performance degradation of 1.28\%.
\textbf{ii) Quantized SNNs.}
Regarding quantized SNNs, QP-SNN, BESTformer, and AGMM exhibit limited performance, whereas our Spikingformer-TaWQ and QKFormer-TaWQ demonstrate a notable advantage in accuracy over these methods. QSDTransformer, which quantizes weights to 4-bit, not only results in larger model sizes than our TaWQ-based models but also consumes significantly more energy. Notably, while QKFormer-TaWQ achieves 2.64\% higher accuracy than QSDTransformer, its energy consumption is only 11.38\%, and its model size is merely 59.56\% of the latter.

\textbf{Comparing with Binary Neural Networks (BNNs).}
We select representative ultra-low-bit (1-bit) quantized BNNs as control groups, including XNOR-Net, Bi-Real Net, AdaBin, and ReActNet, which do not mention the calculation of energy consumption. Spikingformer-TaWQ and QKFormer-TaWQ demonstrate substantial performance improvements over these BNNs. Spikingformer-TaWQ with "8-384" architecture outperforms XNOR-Net, Bi-Real Net, AdaBin, and ReActNet by 21.24\%, 16.04\%, 6.04\%, and 3.04\% in top-1 accuracy, respectively. And QKformer-TaWQ outperforms them by 31.74\%, 26.54\%, 16.54\%, and 13.54\%, respectively.

\begin{figure}[htbp]
    \centering
    \vspace{-0.2cm}
    \includegraphics[width=.475\textwidth]{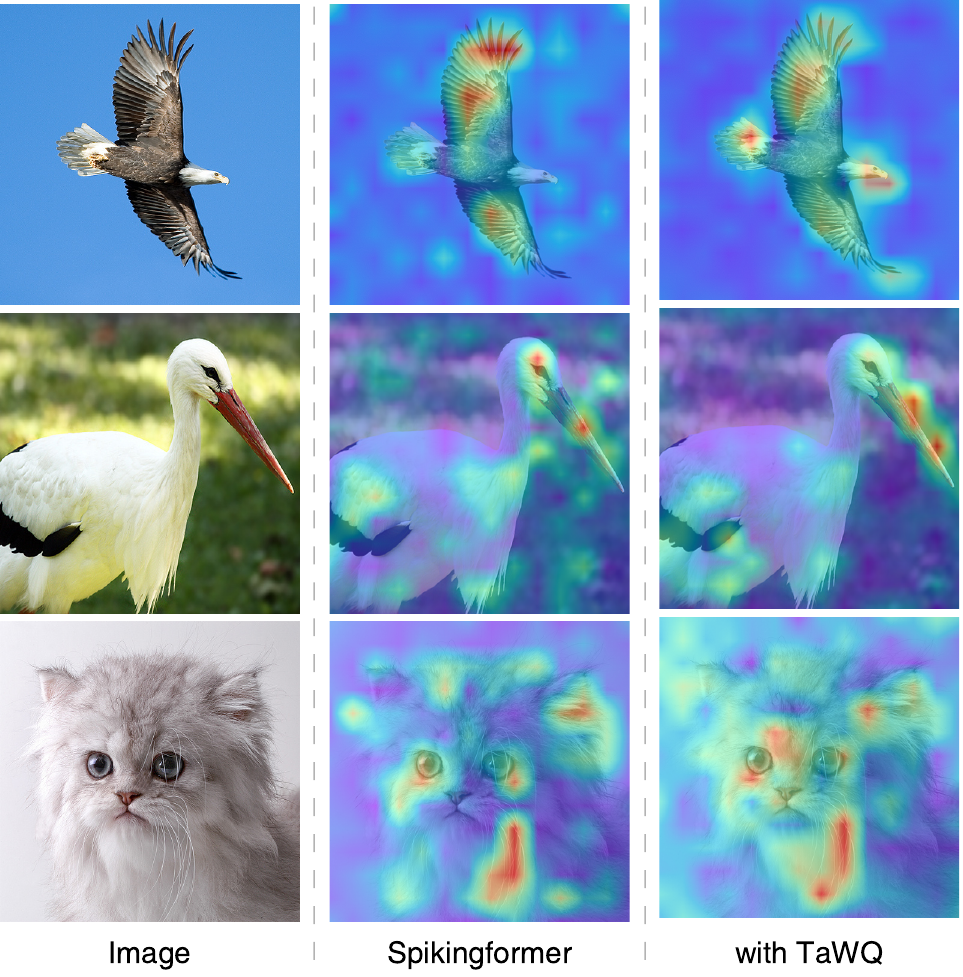}
    \vspace{-0.3cm}
    \caption{Attention maps of the full-precision model and the 1.58-bit quantized model with TaWQ.The images are part of ImageNet's validation set.}
    \label{fig: TaWQ_attn_map_keshihua}
    \vspace{-0.2cm}
\end{figure}

\textbf{Attention Map Visualization.}
We conduct a comparative visualization of attention maps on the validation set for both the full-precision and the TaWQ-quantized model. The attention maps are computed from the last spiking self-attention module of the network and averaged across all timesteps. As shown in Fig. \ref{fig: TaWQ_attn_map_keshihua}, the attention distributions of the models before and after quantization are largely consistent, focusing on the same key features, such as the eagle's wings, the bird's beak, and the cat's whiskers. This indicates that the TaWQ-quantized model retains feature extraction capabilities largely equivalent to those of the full-precision model.

Furthermore, we present additional experimental results in the Supplementary Material, including but not limited to the latency and memory footprint of TaWQ-quantized networks, comparisons with Post-training Quantizations (PTQ), and TaWQ-based quantization results for non-Transformer spiking convolutional networks.

\subsection{Results on CIFAR Classification}
We quantize Spikformer, Spikingformer, and QKFormer on the CIFAR10 and CIFAR100 datasets, resulting in the quantized models Spikformer-TaWQ, Spikingformer-TaWQ, and QKFormer-TaWQ. The results are shown in Table \ref{cifarresults}, \textbf{i) For CIFAR10}, Spikformer-TaWQ and Spikingformer-TaWQ both have a model size of 0.49M (19.0$\times$ smaller), while QKFormer-TaWQ achieves a smaller model size of 0.36M (18.7$\times$ smaller). The accuracy degradation is minor, with drops of only 0.32\%, 0.27\%, and 0.10\%, respectively. Notably, QKFormer-TaWQ, with its compact size, attains an accuracy of 96.08\%, outperforming quantized SNNs such as QP-SNN, Q-SNN, and BESTformer by margins of 0.67\%, 0.54\%, and 0.35\%, respectively. Furthermore, it surpasses the full-precision Spikingformer's accuracy. \textbf{ii) For CIFAR100}, the model sizes of Spikformer-TaWQ and Spikingformer-TaWQ are both 0.53M (17.6$\times$ smaller). Their performance shows increases of 0.51\% and 0.44\%, respectively. QKFormer-TaWQ achieves a model size of 0.39M (17.3$\times$ smaller), with only a 0.30\% accuracy degradation. Notably, Spikingformer-TaWQ attains the highest accuracy of 80.87\%, significantly outperforming other quantized SNNs and surpassing all full-precision models except QKFormer. Additionally, a statistical analysis of the firing rates on CIFAR100 is provided in the Supplementary Material.

\begin{table}[htbp]
  \centering
  \vspace{-0.2cm}
  \caption{Comparison of TaWQ's results on CIFAR10 and CIFAR100. "Bits" denotes the bit-width of weights. The unit of "Size" is "M". "+TaWQ" denotes "xxx-TaWQ", where "xxx" refers to the full-precision model.}
    \resizebox{1.\columnwidth}{!}{
    \begin{tabular}{p{1.55cm}p{0.05cm}<{\centering}p{0.45cm}<{\centering}p{0.3cm}<{\centering}p{1.41cm}<{\centering}p{0.3cm}<{\centering}p{1.41cm}<{\centering}}
    \toprule
    \multicolumn{1}{l}{\multirow{2}[4]{*}{Method}} & \multirow{2}[4]{*}{T} & \multirow{2}[4]{*}{Bits} & \multicolumn{2}{c}{CIFAR10} & \multicolumn{2}{c}{CIFAR100} \\
\cmidrule{4-7}    \multicolumn{1}{l}{} &       &       & Size & Acc(\%) & Size & Acc(\%) \\
    \midrule
    \multicolumn{1}{l}{QP-SNN \cite{qp-snns}} & 2     & 4   & 3.16  & 95.41 & 3.35  & 75.77 \\
    \multicolumn{1}{l}{Q-SNN \cite{q-snns}} & 2     & 1   & 1.62  & 95.54 & -     & 78.82 \\
    \multicolumn{1}{l}{BESTformer \cite{binaryeventdrivenspikingtransformer}} & 4     & 1   & 1.18  & 95.73 & 1.31  & 79.80 \\
    \midrule
    \multicolumn{1}{l}{Spikformer \cite{zhou2023spikformer}} & 4     & 32  & 9.32  & 95.51 & 9.32  & 78.21 \\
    \textbf{\qquad+TaWQ} & 4     & 1.58 & \textbf{0.49}      & \textbf{95.19(-0.32)} & \textbf{0.53}      & \textbf{78.72(+0.51)} \\
    \multicolumn{1}{l}{Spikingformer \cite{cml}} & 4     & 32  & 9.32  & 96.04 & 9.32  & 80.37 \\
    \textbf{\qquad+TaWQ} & 4     & 1.58 & \textbf{0.49}      & \textbf{95.77(-0.27)} & \textbf{0.53}      & \textbf{80.87(+0.44)} \\
    \multicolumn{1}{l}{QKFormer \cite{qkformer}} & 4     & 32  & 6.74  & 96.18 & 6.74  & 81.15 \\
    \textbf{\qquad+TaWQ} & 4     & 1.58 & \textbf{0.36}      & \textbf{96.08(-0.10)} & \textbf{0.39}      & \textbf{80.85(-0.30)} \\
    \bottomrule
    \end{tabular}%
    }
    \label{cifarresults}
    \vspace{-0.4cm}
\end{table}%

\subsection{Results on Neuromorphic Classification}
Following the static CIFAR datasets protocol, we implement 1.58-bit quantization on Spikformer, Spikingformer, and QKFormer by TaWQ. The results are shown in Table \ref{tab:dvsresults}. In terms of model size, the quantized Spikformer-TaWQ and Spikingformer-TaWQ achieve a compact size of 0.14M, representing merely 5.4\% of their full-precision counterparts. QKFormer-TaWQ attains even higher compression with 0.08M, which is only 5.3\% of the full-precision counterpart. Accuracy evaluation reveals minor performance degradation. For CIFAR10-DVS, Spikformer-TaWQ and QKFormer-TaWQ show marginal decreases of 0.7\% and 0.9\% respectively, while Spikingformer-TaWQ demonstrates a 0.3\% improvement. For DVS128-Gesture, Spikformer-TaWQ maintains parity with Spikformer, while Spikingformer-TaWQ and QKFormer-TaWQ exhibit minor accuracy reductions of 0.3\% and 0.7\%, respectively.

\begin{table}[htbp]
  \centering
  \vspace{-0.2cm}
  \caption{Comparison of TaWQ’s results on CIFAR10-DVS and DVS128-Gesture. The unit of "Size" is (M), and "Bits" denotes the bit-width of weights. "+TaWQ" denotes "xxx-TaWQ", where "xxx" refers to the full-precision model.}
    \resizebox{1.\columnwidth}{!}{
    \begin{tabular}{{p{1.55cm}p{0.3cm}<{\centering}p{0.25cm}<{\centering}p{0.45cm}<{\centering}p{1.41cm}<{\centering}p{0.3cm}<{\centering}p{0.25cm}<{\centering}p{0.45cm}<{\centering}p{1.41cm}<{\centering}}}
    \toprule
    \multicolumn{1}{l}{\multirow{2}[4]{*}{Method}} & \multicolumn{4}{c}{CIFAR10-DVS} & \multicolumn{4}{c}{DVS128-Gesture} \\
    \cmidrule{2-9}    \multicolumn{1}{l}{} & Size & T     & Bits  & Acc(\%) & Size & T     & Bits  & Acc(\%) \\
    \midrule
    \multicolumn{1}{l}{QP-SNN \cite{qp-snns}} & 1.61  & 10    & 8   & 82.1  & -     & -     & -     & - \\
    \multicolumn{1}{l}{Q-SNN \cite{q-snns}} & -     & 10    & 1   & 81.6  & -     & 16    & 1   & 97.9 \\
    \multicolumn{1}{l}{BESTformer \cite{binaryeventdrivenspikingtransformer}} & 1.18  & 16    & 1   & 80.8  & -     & -     & -     & - \\
    \midrule
    \multicolumn{1}{l}{Spikformer \cite{zhou2023spikformer}} & 2.57  & 16    & 32  & 80.9  & 2.57  & 16    & 32  & 98.3 \\
    \textbf{\qquad+TaWQ} & \textbf{0.14}      & 16    & 1.58 & \textbf{80.2(-0.7)}      & \textbf{0.14}      & 16    & 1.58 & \textbf{98.3(-0.0)} \\
    \multicolumn{1}{l}{Spikingformer \cite{cml}} & 2.57  & 16    & 32  & 81.4  & 2.57  & 16    & 32  & 98.6 \\
    \textbf{\qquad+TaWQ} & \textbf{0.14}      & 16    & 1.58 & \textbf{81.7(+0.3)}      & \textbf{0.14}      & 16    & 1.58 & \textbf{98.3(-0.3)} \\
    \multicolumn{1}{l}{QKFormer \cite{qkformer}} & 1.50   & 16    & 32  & 84.0    & 1.50   & 16    & 32  & 98.6 \\
    \textbf{\qquad+TaWQ} & \textbf{0.08}      & 16    & 1.58 & \textbf{83.1(-0.9)}      & \textbf{0.08}      & 16    & 1.58 & \textbf{97.9(-0.7)} \\
    \bottomrule
    \end{tabular}%
    }
    \vspace{-0.3cm}
  \label{tab:dvsresults}%
\end{table}%

\subsection{Information Entropy of Quantized Weights}\label{sub:informationentropy}
\begin{figure}[htbp]
    \centering
    \vspace{-0.2cm}
    \includegraphics[width=.325\textwidth]{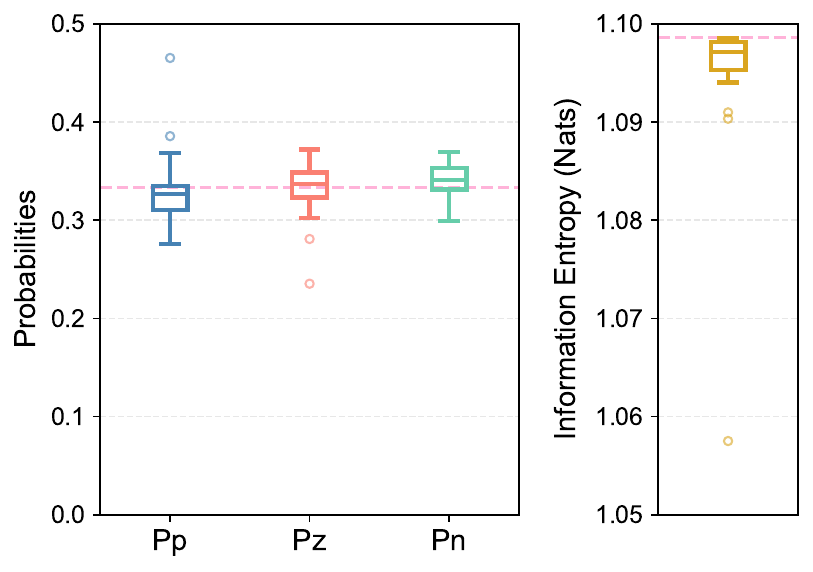}
    \vspace{-0.3cm}
    \caption{Information entropy and weight proportion of TaWQ-quantized QKFormer. 'Pp', 'Pz', and 'Pn' represent the probabilities of +1, 0, and -1 in the weight, respectively. The pink dashed line denotes the optimum.}
    \label{fig: QKFormer-box-ppp-ie}
    \vspace{-0.4cm}
\end{figure}

To assess how TaWQ shapes the weight distribution, we employ information entropy of the quantized weights, with the calculation method detailed in the Supplementary Material. A perfectly balanced ternary weight distribution, equal probabilities for +1, 0, and -1 (i.e., each at $1/3$), yields the theoretical maximum entropy of 1.0986 nats. As summarized in Fig. \ref{fig: QKFormer-box-ppp-ie}, after training QKFormer-TaWQ on CIFAR100, every layer converges to an almost balanced ternary profile: the mean probabilities across all layers are $\mathrm{Pp} = 0.3283$, $\mathrm{Pz} = 0.3320$, and $\mathrm{Pn} = 0.3396$ ( $\mathrm{Pp}$, $\mathrm{Pz}$, and $\mathrm{Pn}$ denote the probabilities of +1, 0, and -1 in the weights, respectively). This corresponds to an average entropy of 1.0952 nats, just 0.0034 nats below the optimum (1.0986), indicating that the network exploits the full expressive capacity of the ternary weights. Notably, although the initial quantized weight distribution under $C_{th}=0.25$ deviates significantly from a uniform distribution, the training procedure drives all the ternary values towards nearly $1/3$, automatically maximizing weight entropy.

\subsection{Ablation Study of Temporal-adaptive Dynamics}\label{ablationstudy}

\begin{figure}[htbp]
    \centering
    \includegraphics[width=.325\textwidth]{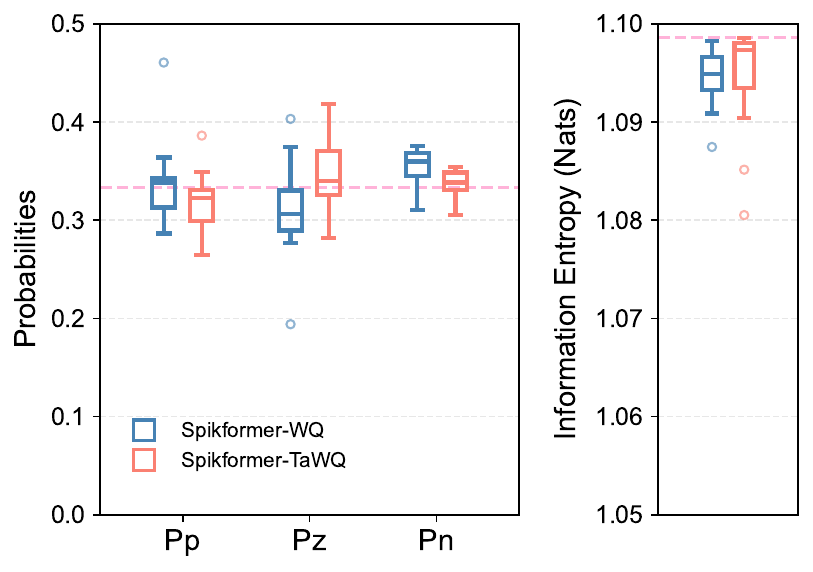}
    \vspace{-0.3cm}
    \caption{Information entropy and weight proportion on static image dataset CIFAR100. The pink dashed line denotes the optimum.}
    \label{fig: ablation-c100-combined-box-ppp-ie}
\end{figure}

\begin{figure}[htbp]
    \centering
    \vspace{-0.1cm}
    \includegraphics[width=.325\textwidth]{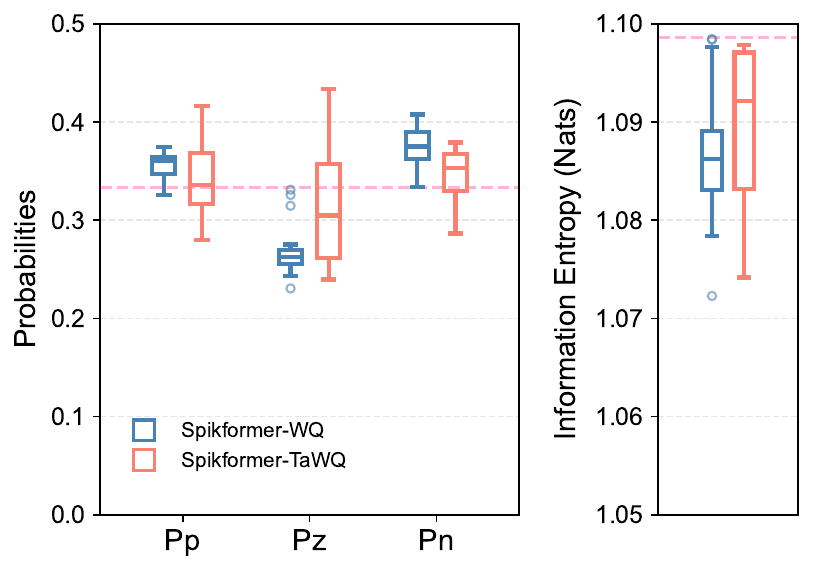}
    \vspace{-0.3cm}
    \caption{Information entropy and weight proportion on neuromorphic dataset DVS128-Gesture. The pink dashed line denotes the optimum.}
    \label{fig: ablation-d128-combined-box-ppp-ie}
    \vspace{-0.3cm}
\end{figure}

We conduct an ablation study on temporal-adaptive dynamics in TaWQ. After removing it, TaWQ degenerates into the $\mathcal{S}(\mathbf{C}_s,C_{th})$ in Eq. (\ref{firetawq}), yielding Spikformer-WQ through quantization of Spikformer. The accuracy of Spikformer-TaWQ and Spikformer-WQ is shown in Table \ref{tab:ablation-acc}. On the DVS128-Gesture dataset, Spikformer-TaWQ outperforms Spikformer-WQ by a margin of 1.1\% in accuracy. Similarly, on CIFAR100, the former exhibited a 0.85\% higher accuracy. The proportion of weights in Spikformer-TaWQ more closely approaches the optimal value of $1/3$ compared to Spikformer-WQ. Consequently, TaWQ yields higher information entropy, as illustrated in  Fig. \ref{fig: ablation-c100-combined-box-ppp-ie} and Fig. \ref{fig: ablation-d128-combined-box-ppp-ie}. These results collectively substantiate the necessity of the temporal-adaptive dynamics. 
Additional ablation studies, including analyses of timesteps and bit-widths, are provided in the Supplementary Material.

\begin{table}[htbp]
  \centering
  \vspace{-0.1cm}
  \caption{Temporal dynamics ablation study results}
    \begin{tabular}{lcc}
    \toprule
    \multicolumn{1}{l}{Dataset} & Method & Acc(\%) \\
    \midrule
    \multirow{2}{*}{DVS128-Gesture} & Spikformer-WQ & 97.2 \\
          & Spikformer-TaWQ & \textbf{98.3(+1.10)} \\
    \multirow{2}{*}{CIFAR100} & Spikformer-WQ & 77.87 \\
          & Spikformer-TaWQ & \textbf{78.72(+0.85)} \\
    \bottomrule
    \end{tabular}%
  \vspace{-0.2cm}
  \label{tab:ablation-acc}%
\end{table}%

\section{Conclusion}
We propose Temporal-adaptive Weight Quantization (TaWQ), an ultra-low-bit weight quantization method inspired by astrocyte-mediated synaptic modulation. By integrating temporal dynamics into the weight quantization process, TaWQ allocates weights temporal-adaptively. Statistics on the learned weight distributions reveal that TaWQ drives the network towards a near-optimal ternary-weight regime, almost fully exploiting the expressiveness of low-bit weights. Extensive experiments on both static and neuromorphic datasets validate that TaWQ narrows the accuracy gap to full-precision models while delivering significant energy savings. The findings underscore TaWQ's potential to guide future neuromorphic algorithms/devices development that enables low-power SNNs deployment in the real world.

\section{Supplementary Material}

\subsection{Efficiency of Biological Nervous Systems}
 We provide quantitative support for our statement about the greater efficiency of biological nervous systems by presenting precise data. The human brain, operating at approximately 20 watts, sustains about $8.6\times 10^6$ neurons and over $1\times 10^{14}$ synapses \cite{humanbrain20w}. SNNs, even when deployed on existing low-power neuromorphic chips (such as TrueNorth \cite{truenorth}, Tianjic \cite{tianjic}, and Loihi2 \cite{loihi2}), demonstrate efficiency levels that remain several orders of magnitude below biological nervous systems. As shown in the Table \ref{tab:human20w}, '\#' represents a number, and '\#Neurons (synapses)/W' denotes the number of neurons or synapses sustained per watt.

\begin{table}[htbp]
  \label{tab:human20w}%
  \centering
  \caption{Quantitative analysis of energy efficiency}
  \scalebox{0.875}{
    \begin{tabular}{p{0.9cm}p{1.0cm}<{\centering}p{1.35cm}<{\centering}p{1.25cm}<{\centering}p{1.35cm}<{\centering}p{1.35cm}<{\centering}}
    \toprule
    Platform & Power(W) & \# Neurons & \#Neurons/W & \#Synapses & \#Synapses/W \\
    \midrule
    TrueNorth & 0.065 & $1.00\times10^6$ & $1.54\times10^7$ & $2.56\times10^6$ & $3.94\times10^7$ \\
    Tianjic & 0.95  & $4.00\times10^4$ & $4.21\times10^4$ & $1.00\times10^7$ & $1.05\times10^7$ \\
    Loihi2 & 1    & $1.00\times10^6$ & $1.00\times10^6$ & $1.20\times10^8$ & $1.20\times10^8$ \\
    Brain & 20   & $\mathbf{8.60\times10^{10}}$ & $\mathbf{4.30\times10^9}$ & $\mathbf{1.00\times10^{14}}$ & $\mathbf{5.00\times10^{12}}$ \\
    \bottomrule
    \end{tabular}%
    }
\end{table}%

\subsection{Information Entropy Computation}\label{iecomputation}

In TaWQ, the quantized weights are 1.58-bit ternary values $\{+1,0,-1\}$, and the probability of each value can be formulated as follows.
\begin{equation}
    f(w_{tri,q}) =\begin{cases}p_{p},\quad w_{tri,q}=+1,\\ p_{z},\quad w_{tri,q}=0,\\ p_{n},\quad w_{tri,q}=-1,\end{cases}
\end{equation}
where $w_{tri,q}$ is a quantized weight, and $p_p+p_z+p_n=1$. We employ information entropy to quantify the information content carried by quantized weights, with the mathematical expression defined as:
\begin{equation}\label{informationentropy}
    \mathcal{H}(w_{tri,q})=-[p_p\mathrm{ln}(p_p)+p_z\mathrm{ln}(p_z)+p_n\mathrm{ln}(p_n)].
\end{equation}
Let $p_p^*,p_z^*,p_n^*=\mathop{\arg\max}\limits_{p_p,p_z,p_n}(\mathcal{H}(w_{tri,q}))$. Analytically, it can be derived that the quantized weights achieve maximum information entropy (1.0986 nats) when $p_p^*=p_z^*=p_n^*=\frac{1}{3}$, satisfying Shannon's entropy maximization principle under a uniform probability distribution.

\subsection{Convergence Analysis of TaWQ Backpropagation}
Since the step function is non-differentiable, we use a surrogate gradient during backpropagation. Boundedness of the weight gradient is a necessary condition for network convergence. We demonstrate that each component of Eq. (7) in the Main Manuscript is bounded.

1) $\frac{\partial L}{\partial\mathbf{W}_{tri,q}[t]}$ is computed from the preceding layer's spiking feature maps. Given finite size and the fact that spiking feature maps contain only values of 0 or 1, so $\frac{\partial L}{\partial\mathbf{W}_{tri,q}[t]}$ is bounded. 

2) $\frac{\partial\mathbf{W}_{tri,q}[t]}{\partial\mathbf{C}_s[t]}<1$, as determined by Eq. (8) in the Main Manuscript. 

3) $\frac{\partial\mathbf{C}_s[t]}{\partial\mathbf{I}_n}=\lambda$. 

4) $\frac{\partial\mathbf{C}_s[t+1]}{\partial\mathbf{C}_s[t]}=1-\lambda$ if $\mathbf{W}_{tri,q}[t]=0$ , otherwise $\frac{\partial\mathbf{C}_s[t+1]}{\partial\mathbf{C}_s[t]}=0$. 

5) $\frac{\partial\mathbf{C}_s[t+1]}{\partial\mathbf{W}_{tri,q}[t]}=(1-\lambda)\times\mathbf{C}_s[t]$ if $\mathbf{W}_{tri,q}[t]<0$, and $\frac{\partial\mathbf{C}_s[t+1]}{\partial\mathbf{W}_{tri,q}[t]}=(\lambda-1)\times\mathbf{C}_s[t]$ if $\mathbf{W}_{tri,q}[t]>0$. 

if $\mathbf{C}_s$ is bounded and the time steps $T$ is finite, then the weight gradient remains bounded. $\mathbf{C}_s$ is derived from $\mathbf{I}$. We initialize
$\mathbf{I}$ using the Kaiming Uniform distribution, ensuring its initial values are bounded. During training, we employ gradient clipping to constrain the magnitude of weight updates. Consequently, after a finite number of training steps, $\mathbf{I}$ remains bounded. This approach guarantees the boundedness of $\mathbf{C}_s$.

Based on the analysis above, we conclude that when the time steps $T$ is finite, the weight gradient remains bounded. This satisfies the necessary condition for network convergence.

\subsection{Consistent Computational Complexity.} 
The computational complexity of the TaWQ is the same as that of the matrix product using time-invariant weights, the latter is employed in existing quantized SNNs \cite{q-snns, binaryeventdrivenspikingtransformer}. Assuming each product between matrix $\mathbf{W}_{tri,q}[t]$ and $\mathbf{X}_i[t]$ has a complexity of $O(N)$, which of the TaWQ is $O(TN)$. If it is a general matrix product, the time-invariant weight $\mathbf{W}$ multiplies temporal inputs $\mathbf{X}_i[t]$ across $T$ timesteps, resulting in a complexity of $O(TN)$, which aligns with the TaWQ. Without introducing additional computational complexity, the TaWQ enhances the representational capacity of quantized SNNs by dynamically assigning distinct weights $\mathbf{W}_{tri,q}[t]$ to $\mathbf{X}_i[t]$, thereby fully exploiting temporal information.

\subsection{Multi-bit TaWQ}
After the 1.58-bit TaWQ, we conduct further in-depth exploration and develop its multi-bit variant, termed mTaWQ (multi-bit Temporal-adaptive Weight Quantization). The dynamics of mTaWQ are defined as follows:
\begin{equation}\label{multibittawq}
\begin{cases}
\mathbf{C}_s\left[t+1\right]=\lambda\cdot\mathbf{C}_s[t](1-\lvert\mathbf{W}_{tri,q}[t]\rvert /n)+(1-\lambda)\cdot\mathbf{I}_{n},\\
\mathbf{W}_{tri,q}[t+1]=\mathcal{Q}(\mathbf{C}_s[t+1],n).
\end{cases}
\end{equation}
Here, $n$ is a positive integer, where $+n$ represents the maximum quantized weight value and $-n$ is the minimum, that is, the quantized weights are constrained to $\{-n,-(n-1),\ldots,0,\ldots,+(n-1),+n\}$. When $n=1$ and $n=2$, the weights are quantized to 1.58-bit and 2.32-bit. With $n=4$, the quantization yields 3.17-bit. And for $n=8$, the weights are quantized to 4.09-bit. $\mathcal{Q}(\mathbf{C}_s,n)$ is the quantization function, and we define it as:
\begin{equation}\label{firemtawq}
    \mathcal{Q}(\mathbf{C}_s,n)=\mathrm{round}(\mathrm{clamp}(\mathbf{C}_s,-n,+n)).
\end{equation}
When $n=1$, $\mathcal{Q}(\mathbf{C}_s,n)$ is functionally equivalent to $\mathcal{S}(\mathbf{C}_s,C_{th})$ with $C_{th}=0.5$. And the surrogate gradient $\mathcal{Q}^{'}(\mathbf{C}_s,n)=1$ if $-n<\mathbf{C}_s<n$, otherwise $\mathcal{Q}^{'}(\mathbf{C}_s,n)=0$.

The ablation study results of bit-width is presented in Section \ref{ablation_study}.

\subsection{Datasets}
We conduct extensive experiments on static image datasets (ImageNet \cite{imagenet}, CIFAR-10/100 \cite{cifar10and100}), neuromorphic datasets (CIFAR10-DVS \cite{cifar10dvs}, DVS128-Gesture \cite{dvs128}) and speech dataset SHD \cite{SHD} to comprehensively evaluate the proposed TaWQ.

ImageNet is the most widely used large-scale benchmark dataset in image classification, comprising 1,000 classes, with a training set of over 1.2 million images and a validation set of 50,000 images. The CIFAR10 dataset contains 50,000 training images and 10,000 test images, while CIFAR100 shares the same total image count (50,000 for training and 10,000 for test) but differs in class granularity: CIFAR10 has 10 classes, whereas CIFAR100 includes 100 classes. Both CIFAR datasets' resolution is 32×32.

CIFAR10-DVS is a neuromorphic dataset converted by sampling static images through a Dynamic Vision Sensor (DVS) camera, containing 10 classes and 10,000 samples with a 9:1 train-test split ratio. DVS128-Gesture is another neuromorphic dataset for gesture recognition, comprising 1,342 samples across 11 gesture classes, collected from 29 individuals under three distinct illumination conditions.

SHD is an audio-based classification benchmark designed for evaluating spiking neural networks. It consists of spoken digits (0-9) in both German and English, covering 20 distinct classes. The dataset contains 10,000 samples, split into 8156 for training and 2264 for testing. The number of channels is 700.

\subsection{Experimental Setups}
The experiments employ LIF neurons with a firing threshold of 1.0, a reset membrane potential of 0, and a membrane time constant of 2.0. These configurations are consistent with many established works, such as Spikformer \cite{zhou2023spikformer}, and represent commonly adopted settings in the field. For experiments on ImageNet, we employ 8 Ascend 910C NPUs, and other datasets are conducted using 1 Ascend 910C NPU.

\textbf{Experimental Setup on ImageNet.}
In this experiment, we employ TaWQ to quantize full-precision models with a batch size of 512 and the AdamW optimizer. The base learning rate ($lr_{\text{base}}$) is set to $6\times 10^{-4}$, and the actual learning rate is calculated as $lr_{\text{base}}\times\text{BatchSize}/256$, resulting in $1.2\times 10^{-3}$, and empoly CosineAnnealing Scheduler. Weight decay is 0.05. The training epochs for Spikingformer-TaWQ and QKFormer-TaWQ were set to 310 and 200, respectively, aligning with their full-precision counterparts. Warmup epochs on ImageNet are 5. Additionally, data augmentation techniques are also aligned with their full-precision counterparts.

\textbf{Experimental Setup on CIFAR.}
The batch size is set to 32, the timesteps are 4, and the learning rate is set to $2\times 10^{-3}$ with AdamW optimizer and CosineAnnealing Scheduler. Weight decay is 0.05. We train models for 400 epochs from scratch, with a warmup for 20 epochs. 

\textbf{Experimental Setup on Neuromorphic Datasets}
Both batch size and timesteps are uniformly set to 16, with learning rates configured as $5\times 10^{-3}$ and $2\times 10^{-3}$ for CIFAR10-DVS and DVS128-Gesture, respectively, and training 106 and 200 epochs from scratch. Optimizer is AdamW with Cosine Annealing Scheduler, the weight decay is 0.06.

\subsection{Energy Consumption}
\subsubsection{Theoretical Energy Consumption}
Comparative analysis of energy consumption in SNNs under the same conditions is critical. Numerous research in the SNN community have adopted energy metrics measured by Horowitz et al. \cite{horowitz20141} on a 45nm hardware platform, where a single 32-bit multiply-accumulate (MAC) operation consumes $E_{MAC}=4.6$ pJ, comprising $3.7$ pJ for multiplication and $E_{AC} = 0.9$ pJ, where the $E_{AC}$ represents the energy consumption of an accumulate (AC) operation. And for a single 8-bit accumulate, $E_{AC} = 0.03$ pJ.

Estimating the energy consumption of SNNs requires first calculating the synaptic operations (SOPs) of neurons, which can be described as follows.
\begin{equation}
    \text{SOPs}^i=\sum_{t=1}^T fr^i_t\times \text{OPs}^i_t,
\end{equation}
where $fr^i_t$ and $\text{OPs}^i_t$ are the firing rate and operations of layer $i$ at timestep $t$, respectively, $T$ is the timesteps in SNNs. OPs comprise binary operations (BOPs) and floating-point operations (FLOPs). Following ReActNet \cite{reactnet}, the OPs is defined as:
\begin{equation}
    \text{OPs}^i_t=\text{BOPs}^i_t/64+\text{FLOPs}^i_t.
\end{equation}

Notably, the weights quantized by TaWQ are 1.58-bit ternary values $\{+1,0,-1\}$. We define a synapse ratio $sr^i_t$ for layer $i$ at timestep $t$, which refers to the proportion of $+1$ (excitatory synapses) and $-1$ (inhibitory synapses) relative to the total weight count. This allows the conversion of ternary operations (TOPs) to equivalent binary operations (BOPs) via the relation $\text{BOPs}^i_t=sr^i_t\times \text{TOPs}^i_t$. In the non-quantized layer, $\text{TOPs}^i_t=0$, leading to $\text{BOPs}^i_t=0$, and $\text{OPs}^i_t=\text{FLOPs}^i_t$. In the TaWQ-quantified layer, $sr^i_t$ is the actual synapse ratio with $\text{FLOPs}^i_t=0$ and $\text{OPs}^i_t=\text{TOPs}^i_t/64\times sr^i_t$.

The TaWQ-quantized SNNs' energy consumption is as follows:
\begin{equation}\label{quant_energy}
E_{quant}=E_{MAC}\cdot\sum_{l=1}^L\mathrm{FLOPs}_{float}^l+E_{AC}\cdot\sum_{n=1}^N\mathrm{SOPs}^n.
\end{equation}
Here, $N$ represents the number of TaWQ-quantized layers, and $L$ is the number of layers with floating-point MAC.

\subsubsection{Energy Consumption with Residual Integer Spikes}
QKFormer adopts the same shortcuts as SEWResNet \cite{sewresnet}, which result in the summation of spikes producing non-binary values (integers $>1$). In the Main Manuscript, operations involving non-binary values and floating-point weights are treated as multiple floating-point additions followed by summation, which is consistent with the original QKFormer paper. However, non-binary values should properly be regarded as floating-point values, and their operations with floating-point weights should be interpreted as floating-point multiplication rather than addition. This makes the $E_{AC}$ inapplicable and necessitates the use of $E_{MAC}$ instead.

We recalculate the energy consumption for QKFormer. A key difference is that TaWQ-quantized weights take values in ${+1, 0, -1}$, the computations between $>1$ integers arising from shortcuts, and these quantized weights still involve additions. Therefore, the use of $E_{AC}$ in the calculation remains necessary. If we treat $>1$ integers as full-precision floating-point values, $E_{AC}=0.9$ pJ. Conversely, if we consider them as 8-bit integers, $E_{AC}$ becomes 0.03 pJ. 

We compute the energy consumption for both scenarios on ImageNet, as shown in the Table \ref{tab:qkformer_energy_re}, where "Power\_add" refers to the original calculation method used in the paper, which interprets $>1$ integers as the sum of multiple spikes. "Power\_int" represents treating $>1$ integers as 8-bit integers. Conversely, "Power\_fp" denotes treating them as full-precision floating-point values. For non-quantized models (where weights remain full-precision), even when $>1$ integers are processed as 8-bit integers, the underlying operations still involve floating-point multiplication. Consequently, only the "Power\_fp" outcome is achievable for non-quantized models in such cases.

The table data reveal that, when interpreting $>1$ integers as 8-bit integers, the full-precision models consume 32.05mJ and 74.12mJ. In contrast, the TaWQ-based models achieve significantly lower energy consumption, at 0.54mJ and 1.52mJ, representing reductions of 98.32\% and 97.95\%, respectively. Even when interpreting $>1$ integers as full-precision floating-point numbers, the TaWQ-based models still exhibit substantial advantages in energy consumption, consuming only 6.00mJ and 24.37mJ. This translates to reductions of 81.28\% and 67.12\%, respectively. Thus, these results conclusively demonstrate that TaWQ effectively and substantially reduces the energy consumption of SNNs.

\begin{table*}[htbp]
  \centering
  \caption{Recalculated Energy Consumption of QKFormer and QKFormer-TaWQ with $T=4$.}
    \begin{tabular}{lccccc}
    \toprule
    Method & Architecture & Bits  & Power\_add(mJ) & Power\_int(mJ) & Power\_fp(mJ) \\
    \midrule
    \multirow{2}{*}{QKFormer \cite{qkformer}} & QKFormer-10-384 & 32    & 15.13 & -     & 32.05 \\
          & QKFormer-10-768 & 32    & 38.91 & -     & 74.12 \\
    \multirow{2}{*}{QKFormer-TaWQ} & QKFormer-10-384 & 1.58  & \textbf{0.45} & \textbf{0.54} & \textbf{6.00} \\
          & QKFormer-10-768 & 1.58  & \textbf{0.99} & \textbf{1.52} & \textbf{24.37} \\
    \bottomrule
    \end{tabular}%
  \label{tab:qkformer_energy_re}%
\end{table*}%

\subsubsection{Energy Consumption with Reading/writing Overhead}
Following \cite{spikeornot}, we have recalculated the energy consumption, with both weights and intermediate states quantized to 8-bit, while spike events are encoded and packed into 8-bit memory words during reading/writing operations (8 spikes per word). We first extend the per-kernel computation described in \cite{spikeornot} to multiple kernels, resulting in $N_{rd}=C_o\times C_i\times k_h\times k_w$, where $N_{rd}$ is the number of weights. In the single-timestep scenario, we quantize weights to 1.58-bit via TaWQ while storing each weight with 2 bits. Extending \cite{spikeornot}'s spike-encoding methodology, the total reading energy becomes $E_{rd_{tot}}=N_{rd}\times(E_{rd}/4+E_{rd}/8)$, which is equal to $E_{wr_{tot}}$ for a writing operation. For the input encoding layer (first layer), where both weights and feature maps remain 8-bit, this yields $E_{rd_{tot}}=N_{rd}\times(E_{rd}/1+E_{rd}/1)$. For non-TaWQ quantized models with 8-bit weights, $E_{rd_{tot}}=N_{rd}\times(E_{rd}/1+E_{rd}/8)$.

We subsequently extend these single-timestep calculations to $T$ timesteps, where weight-reading operations are executed $T$ times. Finally, we incorporate the spatial dimensions of the feature map by including height $H$ and width $W$ in the calculations. Additionally, energy consumption in the neuron is computed consistently using 8-bit precision. The specific energy consumption results for Spikingformer models on ImageNet are presented in the Table \ref{tab:energyhardware}.

It can be observed that the energy consumption of the 8-bit Spikingformer is 0.64mJ, while that of Spikingformer-TaWQ is 0.36mJ, which is just 56.25\% of the former. We additionally calculated the energy consumption specifically for weight reading (Power\_rd) separately. The weight reading energy consumption for Spikingformer is 0.14mJ, while for Spikingformer-TaWQ it is only 34.29\% of the former's consumption, that is 0.048mJ.

In summary, when accounting for both reading/writing energy and internal neuronal computation energy, TaWQ-quantized models demonstrate pronounced energy advantages over their 8-bit counterparts, despite introducing additional temporal dimension during weight quantization.

\begin{table*}[htbp]
  \centering
  \caption{Energy consumption results on hardware with $T=4$.}
    \begin{tabular}{lcccccc}
    \toprule
    Method & Architecture & Bits  & Hardware Bits & Power (mJ) & Power\_rd (mJ) \\
    \midrule
    Spikingformer\cite{cml} & \multirow{2}{*}{Spikingformer-8-768} & 32    & 8     & 0.64  & 0.14 \\
    Spikingformer-TaWQ &       & 1.58  & 2          & \textbf{0.36} & \textbf{0.048} \\
    \bottomrule
    \end{tabular}%
  \label{tab:energyhardware}%
\end{table*}%

\subsection{More Experimental results}

\subsubsection{Statistics of Firing Rates}
\begin{figure}[htbp]
    \centering
    \includegraphics[width=.325\textwidth]{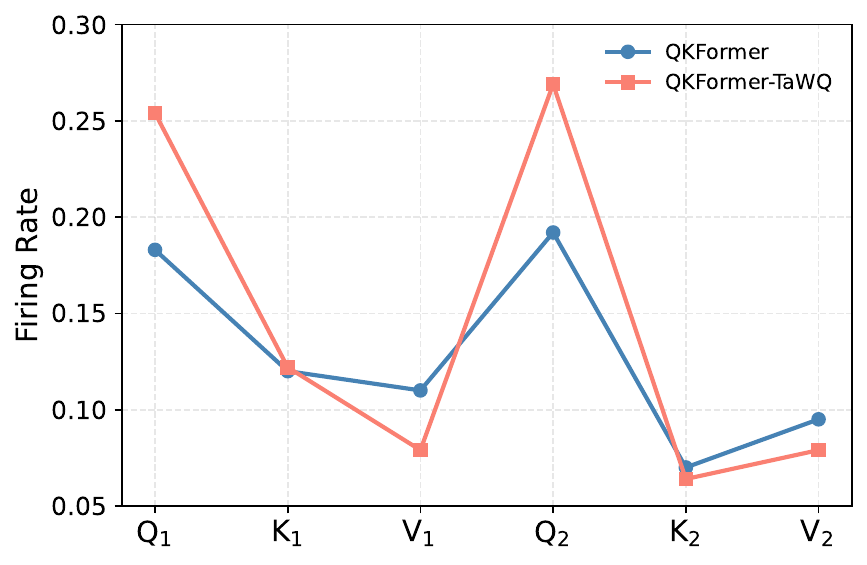}
    \vspace{-0.3cm}
    \caption{Firing rates of full-precision and TaWQ-quantized QKFormer.}
    \label{fig: fr_qkv}
    \vspace{-0.2cm}
\end{figure}

We examine the firing rates of Q, K, and V in the last two spiking self-attention (SSA) modules of both QKFormer and QKFormer-TaWQ trained on CIFAR100, as illustrated in Fig. \ref{fig: fr_qkv}. The QKFormer-TaWQ demonstrates a firing rate trend consistent with the QKFormer, with the highest firing rate in Q, followed by K, and the lowest in V. The Pearson correlation coefficient of the firing rates between QKFormer and QKFormer-TaWQ reaches 0.9733, indicating a strong correlation before and after quantization. Additionally, the mean firing rate after TaWQ quantization is 0.1445, showing a minor deviation from the pre-quantization value of 0.1283. In summary, the TaWQ-quantized model effectively captures the firing rate characteristics of the full-precision model.

\subsubsection{Latency and Memory Footprint}
We further evaluate the latency and memory footprint of the TaWQ quantization method during inference at a single timestep. While the full-precision and quantized model weights are in 32-bit and 1.58-bit, respectively, the inference is conducted on an Ascend 910 NPU with FP16 floating-point precision due to hardware constraints, meaning both the 32-bit and 1.58-bit weights were converted to 16-bit for execution. The batch size is set to 32. The results in Table \ref{tab:latency_memory} indicate that under identical 16-bit conditions, the memory footprint remains the same at 6.88 G, and the latency is nearly identical, with a marginal difference of only 0.2 ms. Significantly, TaWQ-quantized networks exclusively employ addition operations with +1 and -1, therefore, the quantized weights could potentially largely reduce computational demands on hardware specifically designed for low-bit computing.

\begin{table*}[htbp]
  \centering
  \caption{Latency and memory footprint of QKFormer and QKFormer-TaWQ on a single timestep.}
    \begin{tabular}{lcccccc}
    \toprule
    Method & Architecture & Bits  & NPU Bits & Batch & Latency(ms) & Memory(G) \\
    \midrule
    QKFormer \cite{qkformer} & \multirow{2}{*}{QKFormer-10-768} & 32    & \multirow{2}{*}{16} & \multirow{2}{*}{32} & 201.5 & 6.88 \\
    QKFormer-TaWQ &       & 1.58  &       &       & 201.7 & 6.88 \\
    \bottomrule
    \end{tabular}%
  \label{tab:latency_memory}%
\end{table*}%

\subsubsection{Comparisons with Post-training Quantizations}
A comparative analysis is conducted between the 1.58-bit TaWQ-quantized models and their 8-bit quantized counterparts implemented using the Post-Training Quantization (PTQ) method on the ImageNet dataset, with $T=4$. Since Spikingformer and QKFormer only release their largest trained models, we quantized Spikingformer-8-768 and QKFormer-10-768 using TaWQ for comparison. The results are shown in the Table \ref{tab:ptq_tawq}. The results demonstrate that our 1.58-bit Spikingformer-TaWQ achieves 77.42\% accuracy, experiencing only a 0.22\% accuracy drop compared to the full-precision Spikingformer, whereas the 8-bit Spikingformer-PTQ exhibits an accuracy drop of 2.38\%. Meanwhile, QKFormer-10-768-TaWQ achieves 82.94\% accuracy, surpassing the PTQ-quantized 8-bit model by 0.96\%.

\begin{table}[htbp]
  \centering
  \caption{Comparison results of TaWQ-quantized and PTQ (Post-Training Quantization) models.}
    \begin{tabular}{lccc}
    \toprule
    Method & Architecture & Bits  & Acc(\%) \\
    \midrule
    Spikingformer \cite{cml} & \multirow{3}{*}{Spikingformer-8-768} & 32    & 77.64 \\
    Spikingformer-TaWQ &       & \textbf{1.58} & \textbf{77.42(-0.22)} \\
    Spikingformer-PTQ &       & 8     & 75.26(-2.38) \\
    \midrule
    QKFormer \cite{qkformer} & \multirow{3}{*}{QKFormer-10-768} & 32    & 84.22 \\
    QKFormer-TaWQ &       & \textbf{1.58} & \textbf{82.94(-1.28)} \\
    QKFormer-PTQ &       & 8     & 81.98(-2.24) \\
    \bottomrule
    \end{tabular}%
  \label{tab:ptq_tawq}%
\end{table}%

\subsubsection{Results of Non-Transformer Spiking Networks}
We train TaWQ-quantized SEWResNet on the ImageNet dataset using the same epochs, learning rate, batch size, and other hyperparameter settings as those specified in \cite{sewresnet}. We present the result of the 1.58-bit SEW-ResNet18-TaWQ in Table \ref{tab:sew_q}. It achieves a final accuracy of 62.06\% on ImageNet, surpassing the 8-bit QP-SNN by 0.70\%. Compared with the full-precision SEW-ResNet18, this exhibits only a 1.16\% accuracy degradation.

\begin{table}[htbp]
  \centering
  \caption{Results of non-transformer structure on ImageNet.}
    \begin{tabular}{lccc}
    \toprule
    Method & Architecture & Bits  & Acc(\%) \\
    \midrule
    XNOR-Net \cite{xnor-net-eccv2016} & ResNet-18 & 1     & 51.2 \\
    Bi-Real Net \cite{birealnet} & Bi-Real-18 & 1     & 56.4 \\
    QP-SNN \cite{qp-snns} & ResNet-18 & 8     & 61.36 \\
    SEWResNet & \multirow{2}{*}{SEWResNet-18} & 32    & 63.22 \\
    SEWResNet-TaWQ &       & \textbf{1.58} & \textbf{62.06(-1.16)} \\
    \bottomrule
    \end{tabular}%
  \label{tab:sew_q}%
\end{table}%

\subsubsection{Results on SHD classification}
We further validate performance on speech classification tasks using the SHD dataset with a larger timesteps. Input data to the network takes the form $B\times T\times C$, where $B$ is the batch size, $T$ is the temporal dimension, and $C$ is the channel dimension. Following the methodology of \cite{TIM}, the input is processed using spatio-temporal bins, which reduces its dimensionality to $B\times 100\times 140$. Specifically, $T=100$ and $C=140$. We utilize the AdamW optimizer with a Cosine Annealing scheduler. The learning rate is 1e-3, and the batch size is 32. The experimental results on the SHD dataset are presented in Table \ref{tab:shd}, where only minor performance degradation is observed. Specifically, the TaWQ-quantized Spikformer exhibits an accuracy drop of merely 0.35\% compared to its full-precision counterpart, while the TaWQ-quantized Spikingformer shows a reduction of only 0.22\% in accuracy. The first implementation of Spikformer for the SHD dataset is achieved by \cite{TIM}. Our analysis reveals that, compared to this prior work, our quantization method maintains performance advantages with a smaller size.

\begin{table}[htbp]
  \centering
  \caption{Experimental results on the SHD. "*" denotes the self-reimplemented model.}
    \begin{tabular}{lcc}
    \toprule
    Method & Architecture & Acc(\%) \\
    \midrule
    Spikformer \cite{TIM} & \multirow{2}{*}{Spikformer-2-256} & 85.1 \\
    TIM \cite{TIM} &       & 86.3 \\
    \midrule
    Spikformer \cite{zhou2023spikformer} & \multirow{2}{*}{Spikformer-1-128*} & 91.16 \\
    Spikformer-TaWQ &       & \textbf{90.81(-0.35)} \\
    \midrule
    Spikingformer \cite{cml} & \multirow{2}{*}{Spikingformer-1-128*} & 91.74 \\
    Spikingformer-TaWQ &       & \textbf{91.52(-0.22)} \\
    \bottomrule
    \end{tabular}%
  \label{tab:shd}%
\end{table}%

\subsubsection{Ablation Study} \label{ablation_study}
\textbf{Ablation Study of Bit-width.}
We conduct an ablation study into multi-bit TaWQ (mTaWQ) using the CIFAR100 and QKFormer-mTaWQ, with the results shown in Table \ref{tab:mtawq_results}. The 1.58-bit model is quantized using TaWQ, while the higher-bit model is obtained via mTaWQ. 
The model size decreases as the bit-width reduces, while the accuracy does not show a significant decline. 
When quantized to 4.09-bit, the model size is 0.91M, achieving an accuracy of 80.97\%.
At 1.58-bit, the model size is reduced to 42.86\% of the 4.09-bit model, with only a 0.12\% decrease in accuracy, indicating that the 1.58-bit weights already encapsulate sufficient information.

\begin{table}[htbp]
  \centering
  \vspace{-0.1cm}
  \caption{Bit-width ablation study results.}
    \begin{tabular}{ccccc}
    \toprule
    Method   & Bits & T     & Size(M) & Acc(\%) \\
    \midrule
    \multirow{3}{*}{mTaWQ}
                 & 4.09 & 4     & 0.91      & 80.97 \\
                 & 3.17 & 4     & 0.72      & 80.61 \\
                 & 2.32 & 4     & 0.55      & 80.85 \\
    \midrule
     TaWQ      & 1.58 & 4     & 0.39      & 80.85 \\
    \bottomrule
    \end{tabular}%
  \label{tab:mtawq_results}
  \vspace{0.1cm}
\end{table}%

\begin{figure}[htbp]
    \centering
    \includegraphics[width=.325\textwidth]{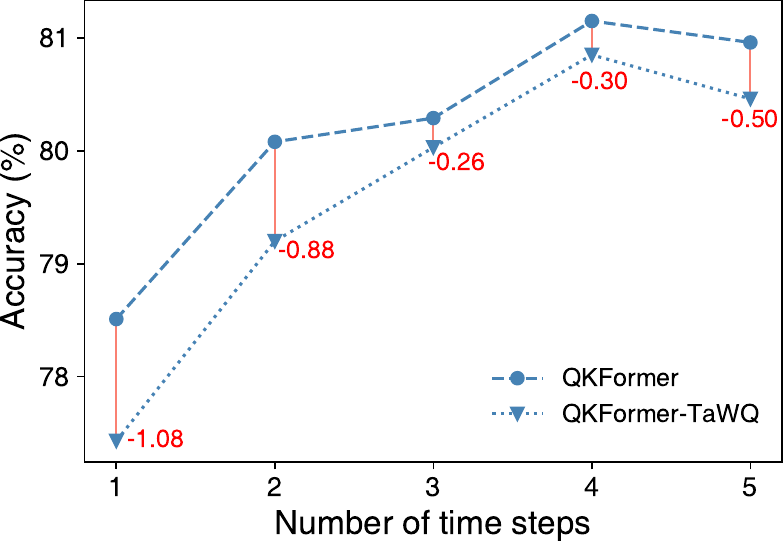}
    \vspace{-0.2cm}
    \caption{Timesteps ablation study results of TaWQ-quantized QKFormer.}
    \label{fig: Tablation}
\end{figure}

\textbf{Ablation Study of Timesteps.}
The performance of SNNs shows a strong dependence on timestep configurations. We conduct a timestep ablation study on the CIFAR-100 dataset. As shown in Fig. \ref{fig: Tablation}, the accuracy of QKFormer exhibits an increasing trend with timestep progression, peaking at $T=4$. Notably, the accuracy of the quantized QKFormer-TaWQ follows an identical trend to QKFormer, indicating that the performance profile remains consistent despite TaWQ's application. By analyzing the accuracy difference, we observe that QKFormer-TaWQ achieves the closest accuracy with QKFormer at $T=3$, with a marginal gap of 0.26\%, followed by $T=4$ with a marginal gap of 0.30\%. Overall, selecting T=4 achieves high quantization performance while maintaining low degradation.

\textbf{Ablation Study of Quantization Threshold.}
The threshold parameter $C_{th}$ in TaWQ affects the performance of quantized SNNs. We conduct $C_{th}$ ablation experiments on the CIFAR-100 dataset using the QKFormer-TaWQ, and the experimental results are shown in Table \ref{tab:Wthablation}. We keep the timestep fixed at $T=4$ and weights quantized to 1.58-bit, only by varying the value of $C_{th}$, it can be observed that the highest accuracy of 80.85\% is achieved at $C_{th}=0.25$, followed by 80.43\% ($C_{th}=0.15$) and 80.40\% ($C_{th}=0.35$), the accuracy further decreases when $C_{th}=0.50$, yields 80.34\%. These results indicate that $C_{th}=0.25$ represents the closest optimal value for QKFormer-TaWQ among these $C_{th}$ on the CIFAR-100 benchmark.

\begin{table}[htbp]
  \centering
  \caption{The $C_{th}$ ablation study results.}
    \begin{tabular}{cccc}
    \toprule
    $C_{th}$   & Bits(W-A) & T     & Acc(\%) \\
    \midrule
    0.15  & \multirow{4}{*}{1.58-1} & 4     & 80.43 \\
    0.25  &       & 4     & 80.85 \\
    0.35  &       & 4     & 80.40 \\
    0.50   &       & 4     & 80.34 \\
    \bottomrule
    \end{tabular}%
  \label{tab:Wthablation}%
\end{table}%

\subsection{Future work}
Our future work will research TaWQ-based quantization in diverse tasks such as detection, segmentation, and language. Additionally, multi-bit quantization variants of TaWQ will be extended to large-scale models. Furthermore, TaWQ-quantized models will be deployed on hardware platforms, including neuromorphic chips and Field Programmable Gate Arrays (FPGAs), to evaluate actual energy consumption and performance under real-world situations.

\vfill

\end{document}